\pdfoutput=1

\documentclass[11pt]{article}

\usepackage[final]{acl}
\usepackage{acl}

\usepackage{times}
\usepackage{latexsym}
\usepackage{svg}
\usepackage[T1]{fontenc}

\usepackage[utf8]{inputenc}

\usepackage{microtype}

\usepackage{inconsolata}

\usepackage{booktabs}
\usepackage{multirow}
\usepackage{float}
\usepackage{graphicx}
\usepackage{amsmath}
\usepackage{amsfonts}
\usepackage{url}
\usepackage{bm}
\usepackage{xspace}
\usepackage[most]{tcolorbox}
\usepackage{adjustbox}
\usepackage{colortbl}

\usepackage{tabularx}
\usepackage{fdsymbol}

\usepackage{xcolor}
\usepackage{geometry}
\usepackage{array}
\usepackage{booktabs}

\definecolor{contextHighlight}{RGB}{173,216,230} 
\definecolor{cotHighlight}{RGB}{255,204,153}     
\definecolor{coeHighlight}{RGB}{204,255,204}     

\usepackage{cleveref}

\usepackage{pifont}
\newcommand{\cmark}{\textcolor{blue}{\ding{51}}}%
\newcommand{\xmark}{\textcolor{orange}{\ding{55}}}

\usepackage{color, colortbl}
\definecolor{LightCyan}{rgb}{0.88,1,1}

\definecolor{questionblue}{RGB}{0,0,180}      
\definecolor{answergreen}{RGB}{0,128,0}         
\definecolor{cotred}{RGB}{200,0,0}              
\definecolor{coeshortorange}{RGB}{255,102,0}    
\definecolor{headergray}{RGB}{211,211,211}       

\definecolor{highlightblue}{RGB}{173,216,230}

\newcommand{\minisection}[1]{\noindent{\textbf{#1}}}
\newcommand{\tool}{\textsc{E2G}~}
\newcommand{\toolnospace}{\textsc{E2G}}

\newcommand{\toolnew}{\textsc{CoE}~}
\newcommand{\toolnewnospace}{\textsc{CoE}}
\newcommand{\CoEshort}{\textsc{CoE-short}}
\newcommand{\CoElong}{\textsc{CoE-long}} 
\title{Chain of Evidences and Evidence to Generate: Prompting for Context Grounded and Retrieval Augmented Reasoning
}

\author{Md Rizwan Parvez 
\\ Qatar Computing Research Institute (QCRI)\\
\texttt{mparvez@hbku.edu.qa}}

\begin{document}
\maketitle

\begin{abstract}

While chain-of-thoughts (CoT) prompting has revolutionized how LLMs perform reasoning tasks, its current methods and variations (e.g, Self-consistency, ReACT, Reflexion, Tree-of-Thoughts (ToT), Cumulative Reasoning (CR) etc.,) suffer from limitations like limited context grounding, hallucination/inconsistent output generation, and iterative sluggishness. To overcome these challenges, we introduce a novel mono/dual-step zero-shot prompting framework built upon two unique strategies \textbf{Chain of Evidences (\toolnewnospace)} and \textbf{Evidence to Generate (\toolnospace)}. Instead of unverified reasoning claims, our innovative approaches leverage the power of "evidence for decision making" by first focusing exclusively on the thought sequences explicitly mentioned in the context which then serve as extracted evidence, guiding the LLM's output generation process with greater precision and efficiency. This simple yet potent approach unlocks the full potential of chain-of-thoughts prompting, facilitating faster, more reliable, and contextually aware reasoning in LLMs. Our framework consistently achieves remarkable results across various knowledge-intensive reasoning and generation tasks, surpassing baseline approaches with state-of-the-art LLMs. For instance, (i) on the LogiQA benchmark using GPT-4, \toolnew achieves a new state-of-the-art accuracy of 53.8\%, surpassing CoT by 18\%, ToT by 11\%, and CR by 9\%; (ii) \toolnew with PaLM-2 outperforms the variable-shot performance of Gemini Ultra by 0.9 F1 points, achieving an F1 score of 83.3 on DROP. We release our prompts and outputs on these benchmarks as a new instruction tuning dataset for future research at \emph{Hugging Face}\footnote{\url{https://huggingface.co/datasets/kagnlp/Chain-of-Evidences/}}.


\end{abstract}

\section{Introduction}
\label{sec:intro}


\begin{figure}[t]
\vspace{-2mm}
    \centering
    \includegraphics[width=0.47\textwidth]{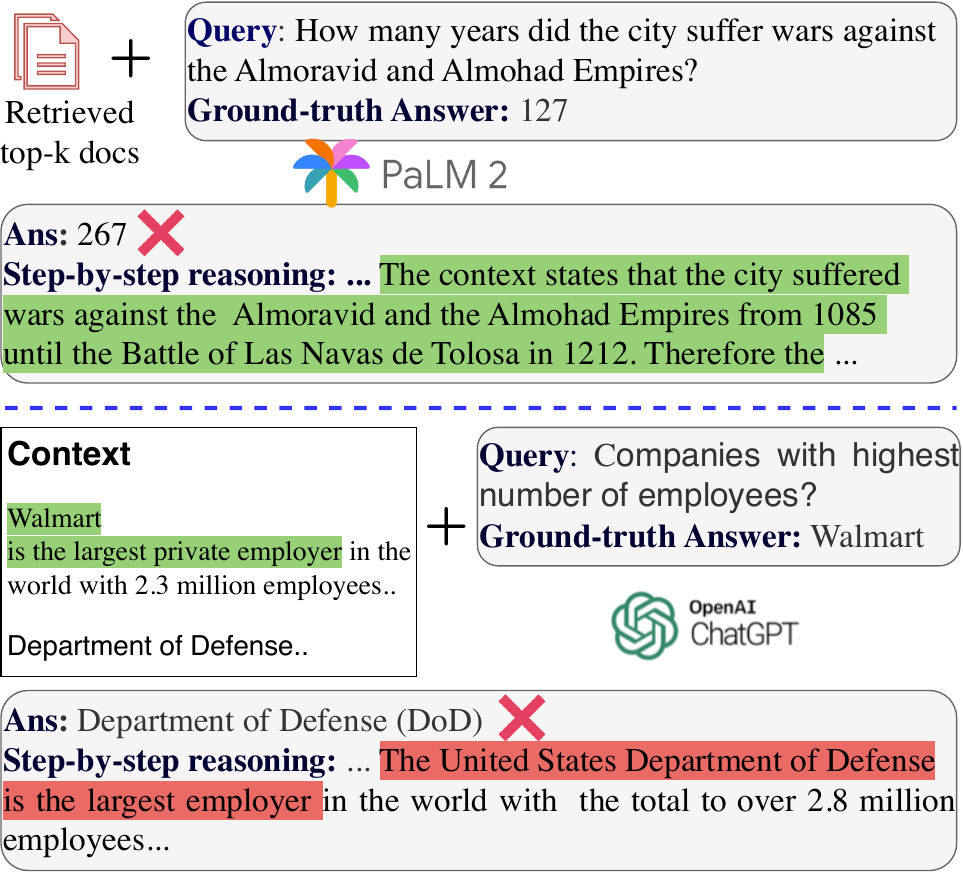}
    \caption{CoT \& variants falter in context-aware reasoning. Top: Overwhelming long-text complexity leads models' failure even when it generates partially/fully correct reasoning (in green). Bottom: Ungrounded internal reasoning fails to grasp context, confusing "DoD" (ungroundeded private org in red) vs Walmart (in green). }
\vspace{-3mm}
\label{fig:cot_error}
\end{figure}


\begin{figure*}[ht]
\vspace{-10pt}
    \centering
    \includegraphics[width=0.9\textwidth]{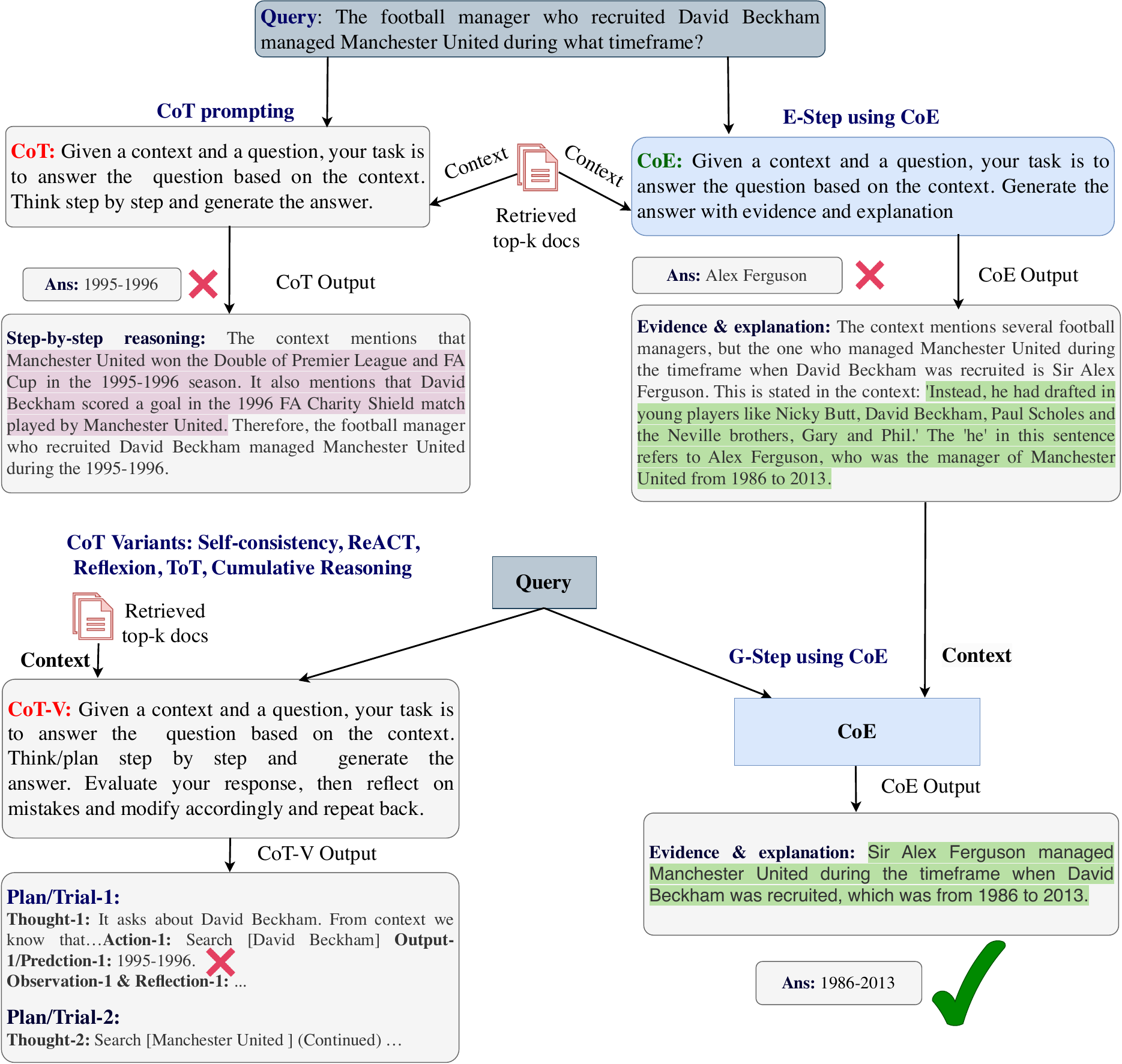}
    \vspace{-1mm}
    \caption{(left) CoT and generic view of its (iterative) variants, (right) The \textsc{E2G} pipeline: In E-step our "generate ans with evidence and explanation" instruction extracts the rationales, coupled with the ans, grounded in the original context, then in G step we use the same instruction to derive the final answer solely from the "evidence and explanation"  or along with the original context. }
\label{fig:e2g-pipeline}
\end{figure*}

Retrieval-augmented or context-based generation serves as a mean for leveraging relevant information, empowering large language models (LLMs) to reduce the factual errors in their generation \citep{islam-etal-2024-open, asai2023retrieval, asai2023self}. However, despite the expansion in model and data size, LLMs struggle in contextual reasoning. This challenge is further amplified when dealing with retrieved information that are often long and imperfect text with distractive contents.

To bolster LLM's reasoning capabilities, the Chain-of-Thought (CoT) prompting paradigm has emerged as a potent tool \citep{cot}. Subsequent methods, including Self-consistency (SC; \citep{wang2022self}), ReACT \citep{yao2022react}, Reflexion \citep{shinn2023reflexion}, Tree of Thoughts (ToT; \citep{yao2023tree}), and Cumulative Reasoning (CR; \citep{zhang2023cumulative}), generalize CoT with various 
multi-objective, ensemble-based, or tool-augmented, and trial \& error 
approaches 
but
do not address the complexities of context-grounded or retrieval augmented generations (RAG). We highlight two of their pivotal bottlenecks: (i) CoT focuses solely on expanding steps without verifying hypotheses; (ii) excessively long retrieved text can lead to incorrect conclusions even with valid CoT reasonings (example in Figure \ref{fig:cot_error}).



Multi-step reasoning prompting has emerged as a promising alternative to traditional chain-of-thought (CoT) approaches by decomposing complex problems into sequential reasoning substeps \cite{dhuliawala2023chain, wang2023boosting-cok, zhao2023verify, trivedi-etal-2023-interleaving, fu2022complexity, creswell2022selection, li2023chainofknowledge}. However, these techniques typically require rigorous verification of each intermediate step. Although simpler iterative verification strategies—such as self-check \cite{miao2023selfcheck} and self-refine \cite{madaan2024self}—have been proposed, they do not fully address the challenges inherent in long-context processing or retrieval-augmented generation.
Moreover, they often rely on disparate intermediate prompts—such as rationale selection and inference/premise derivation—that necessitate k-shot annotated in-context exemplars, which are often difficult to construct \cite{islam2025codesim, islam-etal-2024-mapcoder, yasunaga2024large}.
Therefore, unlocking CoT's true potential for RAG \&  context driven reasoning remains unanswered. To address, in this paper, we propose a simple verification-free zero-shot prompting framework for context-grounded and retrieval augmented reasoning.   

Our framework consists of two unique and real-time prompting strategies particularly tailored for long context reasoning.  First, single-step \textbf{Chain-of-Evidences (\toolnewnospace):} to address the problem of ungrounded reasoning hypotheses, our designed prompt asks for specific thought sequences that are explicitly mentioned in the context.  We call this series of intermediate reasoning steps, with directly extracted rationales from the given context, '\emph{Evidence}' (as in human decision making). Our key distinction from existing CoT approaches is that instead of mere "thinking step-by-step" \citep{kojima2022large} our prompt instruction asks for "step-by-step reasoning with explicit evidence and explanation". 

Second, dual-step \textbf{Evidence to Generate (\toolnospace):} to facilitate LLMs' answering the query properly even with retrieval augmented long-text contexts, we split the task into steps. In the first step (E), we adopt prompts similar to \toolnew and generate both the \emph{Answer} \& \texttt{$Evidence$} . Then in next step (G), we pass only the \texttt{$Evidence$} as context for a second round of \toolnew to LLM. G Step \emph{Answer} is predicted as the final answer. In contrast to complex long original context in E step, the \texttt{$Evidence$} is a concise short text that directly answer the input query, G step is very fast, and simpler for the model to generate answer. 

In experiments with different LLMs,  we show that our prompts consistently outperform existing approaches in a diverse set of eight context-driven tasks, including natural QA, complex multi-hop, long-form QA, fact checking, dialog generation, and reading comprehension tasks. Since, even with such techniques, it is non-trivial to comprehend why and how this works and how to setup the prompt to function correctly, cost-effectively, and robustly. To this end, we perform case studies, analyze different alternatives and reveal the strengths and weaknesses of our approach.
We open-source our prompts and outputs on these benchmarks as a new instruction tuning dataset for future research.



\section{Related Works and Preliminaries}
\label{sec:related-work}

\subsection{Prompting LLMs}
\label{sec:cot}
Various prompting paradigms have been studied in literature toward enhancing reasoning in LLMs. In Section \ref{sec:intro}, we provide a (non-exhaustive) list of CoT approaches.  Among others,  search-based \citep{pryzant2023automatic, lu2021neurologic}, Program-aided LLM generation \citep{liu2023llm, gao2023pal, jung2022maieutic, zhu2022solving}, self generation of prompts \citep{he2023exploring, yasunaga2023large, sun2022recitation, kim2022self, li2022self}, self evaluation based approaches \citep{madaan2023self, xie2023decomposition, kim2023language, paul2023refiner} have been studied. 
Other works have also been extended with more complex multi-step reasoning procedure (e.g., using a different fine-tuned model \citep{yu2023chain-of-note, nye2021show, lester-etal-2021-power}) or for domain specific applications \citep{parvez-etal-2023-retrieval, parvez-etal-2021-retrieval-augmented, ouyang2022training, sanh2021multitask, wei2021finetuned}. 

\subsection{Chain-of-Thoughts (CoT) Prompting}
\label{sec:cot}
Chain-of-thoughts (CoT; \citep{cot}) is a prompting framework that guides LLMs to produce intermediate reasoning steps towards the final answer, enhancing its reasoning. Original version of CoT employs a few-shot version by providing multiple exemplars of the reasoning process (question–reasoning–answer), leveraging LLMs’ in-context learning abilities. However, due to the  requirement of labeled exemplars, it quickly evolved with a 0-shot instance \citep{kojima2022large}. 0-shot CoT prompts LLMs with a general instruction like “think step by step” to produce intermediate reasoning steps (See {{ Figure \ref{fig:e2g-pipeline}).}} 


\renewcommand{\arraystretch}{1.3}
\begin{table*}[t]
\scriptsize
    \centering
    \resizebox{0.9\linewidth}{!}{
    \begin{tabular}{c|c|c|c|c|c|c|c}
        \hline
        \rowcolor{gray!25}
        \textbf{|Context|} & \textbf{Multi-} & \textbf{Context-} & \textbf{Cost-} & \multicolumn{2}{c|}{\textbf{E-step}} & \multicolumn{2}{c}{\textbf{G-step}} \\ 
        \cline{5-8}
        \rowcolor{gray!25}
        \textbf{>200} & \textbf{Query} & \textbf{Aware} & \textbf{Minimize} & \textbf{Prompt} & \textbf{Context} & \textbf{Prompt} & \textbf{Context} \\
        \hline
        \rowcolor{white}\xmark & \xmark & \xmark & \xmark & CoE-Long & - & - & - \\ 
        \rowcolor{gray!15}\xmark & \xmark & \xmark & \cmark & CoE-Short & - & - & - \\ 
        \rowcolor{white}\xmark & \xmark & \cmark & \xmark & CoE-Long & OC & - & - \\ 
        \rowcolor{gray!15}\xmark & \xmark & \cmark & \cmark & CoE-Short & OC & - & - \\ 
        \rowcolor{white}\xmark & \cmark & \xmark & \xmark & CoE-Long & - & - & - \\ 
        \rowcolor{gray!15}\xmark & \cmark & \xmark & \cmark & CoE-Short & - & - & - \\ 
        \rowcolor{white}\xmark & \cmark & \cmark & \xmark & CoE-Long & OC & CoE-Long & E + OC \\ 
        \rowcolor{gray!15}\xmark & \cmark & \cmark & \cmark & CoE-Short & OC & CoE-Short & E + OC \\ 
        \rowcolor{white}\cmark & \xmark & \cmark & \xmark & CoE-Long & OC & CoE-Long & E \\ 
        \rowcolor{gray!15}\cmark & \xmark & \cmark & \cmark & CoE-Short & OC & CoE-Short & E \\ 
        \rowcolor{white}\cmark & \cmark & \cmark & \xmark & CoE-Long & OC & CoE-Long & E + OC \\ 
        \rowcolor{gray!15}\cmark & \cmark & \cmark & \cmark & CoE-Short & OC & CoE-Short & E + OC \\ 
        \hline
    \end{tabular}
    }
   \caption{Recommended alternative mono/2-step prompts, \& contexts in each step. OC, E refer to original context, \texttt{$Evidence$}.}
    \label{tab:bp}
\end{table*}

\section{Our Prompting Framework}
\label{sec:framework}
In this section, we develop our prompting framework for context-grounding and retrieval augmented long-text reasoning. We design two unique (mono/dual-step) prompts that does not require any exemplars and removes the hurdles of choosing multi-objective instructions. Below we first present the prompt instruction for defining the objective for the target task (a.k.a system prompt), next the single-step prompting technique  \textbf{Chain of Evidences (\toolnewnospace)} and finally dual-step \textbf{Evidence to Generate (\toolnospace)} that uses \toolnew twice.


\subsection{System/Objective Instruction }
\label{sec:agent}
Our proposed framework is a single-intent system, having only one target task to solve at a time. Given a target task \texttt{T}, our objective/system prompt is:


\begin{tcolorbox}[colback=blue!5!white, colframe=blue!25!black, fonttitle=\footnotesize, boxrule=0.2mm, sharp corners]
\texttt{\# You are a/an [\texttt{T}] agent. Given a context and a [\texttt{T[x]}] as input, please give a [\texttt{T[y]}]   output based on the context.
}
\end{tcolorbox}

\texttt{T[x]} and \texttt{T[y]} depends on the task \texttt{T}. Examples of \texttt{T}, \texttt{T[x]} and \texttt{T[y]} are (QA, fact verification, dialogue generation), (question, claim, previous dialogue), and (answer, judgement, next turn dialogue) respectively. An example for fact checking:


\begin{tcolorbox}[colback=blue!5!white, colframe=blue!25!black, fonttitle=\footnotesize, boxrule=0.2mm, sharp corners]
\texttt{\#  You are a text classification agent. Given a context and a claim, please give a judgement to the claim ('SUPPORTS' or 'REFUTES') based on the context.}
\end{tcolorbox}

\subsection{Chain of Evidences (\toolnewnospace)}
\label{sec:coe}
While the 0-shot CoT instruction (i.e., Answer the question. Think step-by-step.) expands the query answer generation into small reasoning steps, it does not focus on context-grounding and generate imaginary hypotheses. To address, our prompt asks for answering the query specifically with evidence and explanation from context. We design two alternatives \CoEshort~ \& \CoElong.


\begin{tcolorbox}[title= CoE-Short, colback=blue!5!white, colframe=blue!25!black, fonttitle=\footnotesize, boxrule=0.2mm, sharp corners]
\texttt{\# Objective Instruction from Section \ref{sec:agent} \\}
\texttt{\# Generate the answer with evidence and explanation.}
\end{tcolorbox}


\begin{tcolorbox}[title= CoE-Long, colback=blue!5!white, colframe=blue!25!black, fonttitle=\footnotesize, boxrule=0.2mm, sharp corners]
\texttt{\# Objective Instruction from Section \ref{sec:agent}\\}
\texttt{\# Think step-by-step and generate the answer with evidence and explanation.}
\end{tcolorbox}

 An overview is in Figure \ref{fig:e2g-pipeline}. However, depending on the task \texttt{T}, we add one or two additional instructions to clarify how the answer should be generated, and what should be the output format: 


\begin{tcolorbox}[colback=blue!5!white, colframe=blue!25!black, fonttitle=\footnotesize, boxrule=0.2mm, sharp corners]
\texttt{\# Your answer must be the either of ('SUPPORTS' or 'REFUTES') based on the claim and the context. \\} 
\texttt{\# Generate your response in a json output format with an 'answer' tag and an 'evidence and explanation' tag }
\end{tcolorbox}

While both \toolnew prompts generates more context-driven reasonings which are often very concise w.r.t the original context, \CoElong~  prompt, which includes "step-by-step" command, instructs the model to generate more verbose and expanded reasoning paths in compare to \CoEshort. Hence, typically \CoElong~  tends to be more accurate (e.g., for commonsense, multi-step reasoning, or arithmetic cases) while \CoEshort~  is more cost-effective.

\begin{table*}[t]
\centering
\scriptsize 
    \resizebox{\linewidth}{!}{
        \begin{tabular}{l|c|c|c|c|c}
            \hline
            Dataset & Size & Reasoning & $|$Context$|$ & Task  & Metric   \\
            \hline 
             LogiQA  & 651 & \multirow{2}{*}{MRC} & 77 & Logical Reasoning & Acc 	 \\
              DROP  & 500 &  & 196 & Arithmetic Reasoning & F1 	 \\
              \hline
               HotpotQA  & 7.41K$^{CG}$/1.5K$^P$ & {Distarctor}  & 1106 & Multi-hop QA & \multirow{3}{*}{EM, F1 }	 \\
            \cline{0-4}
           NQ  & 500 & \multirow{5}{*}{RAG} & \multirow{5}{*}{650-675} & \multirow{2}{*}{Open-domain QA } &  	 \\
            TQA & 1.5K & & &  &  \\
             \cline{5-6}
            WOW & 500 & & & Know. Grounded Dialouge Gen. &  \multirow{2}{*}{ F1 }  \\
            ELI5 & 300 & & & Long Form QA &  \\
           \cline{5-6}
            FEVER& 10.1K$^{CG}$/.1K$^{P}$ & &  & Fact Verification    & Acc\\
            
            \hline
        \end{tabular}
        }
        \vspace{-10pt}
        \caption{Evaluation Datasets.  MRC, and distractor denote machine reading comprehension, and context with distracting documents. 
        |Context| denotes avg token length. $^{CG/P}$ denotes with ChatGPT and  PALM-2 respectively. }
        \label{tab:dataset-stats}
    \vspace{-2mm}
\end{table*}

\subsection{Adaptation}
\label{sec:best-parctice}

In this section, we outline how our framework adapts to various tasks and objectives. Our framework offers choices between mono/dual step prompting, \toolnew alternatives, and context inputs. Considering task complexity, we examine the nature of the task (context-aware or context-free), context length, and query complexity (single or multi-question). Regarding objectives, we prioritize cost optimization or performance triggering. Our design principles are mainly three-folds:
\begin{enumerate}
    \itemsep0em 
    \item Single-step \toolnew is generally sufficient, except for longer contexts where \toolnospace is employed.
\item Cost-effectiveness is tied to the number of steps or LLM API calls. Thus, for \toolnospace, \CoEshort~  is more cost-effective in each step, while \CoElong~  offers granular reasoning steps, enhancing performance, particularly in context-less reasoning tasks like arithmetic and commonsense.
\item The G-step context is typically derived from \texttt{$Evidence$} from the E-step. However, for queries involving multiple sub-queries or answers, a brief \texttt{$Evidence$} may provide only partial answers. In such cases, the G-step context should include \texttt{$Evidence$} concatenated with the original context. \Cref{tab:bp} summarizes these principles.
\end{enumerate}
Another objective, we consider is inference time. While the worst-case runtime of our approach is approximately double that of CoT, shorter \texttt{$Evidence$} reduces runtime (e.g., 1.5s vs CoT's 1s on average), making it suitable for practical use cases. However, more constrained inference time can be achieved via single-step \toolnewnospace.

\section{Experimental Setup}
\label{sec:exp-data}

We evaluate our prompting framework across eight context-intensive language tasks, requiring reasoning over given contexts, including those with distracting documents and retrieval augmentation for generation. Using three LLMs (ChatGPT, GPT-4, PaLM-2 (540B)) via APIs, we conduct comprehensive experiments. 
Due to the size of the datasets, we use sampling and dev splits for evaluation, following established practices. We compare our results with CoT baselines and other frameworks from the literature, reproducing 0-shot CoT where necessary. For retrieval tasks, we utilize datasets from \citet{filco}, comprising DPR \citep{karpukhin-etal-2020-dense}  retrieved top-5 context documents from Wikipedia. Benchmark summaries are in Table \ref{tab:dataset-stats}. By default, we use the single-step \CoElong~ for LogiQA \& DROP, and two-step \tool (with \CoEshort) for other tasks 
where G-step contexts are sourced from \texttt{$Evidence$}, unless otherwise specified. We use \citet{dalvi2023llmebench} in implementation.

\section{Main Results}

\label{sec:exp}


\minisection{Arithmetic/Logical Context Reasoning}
\label{sec:logiQA}
We evaluate our approach on the MRC tasks LogiQA and DROP, known for heavy arithmetic and logical reasoning complexities. LogiQA tasks involve choosing among four options inferred from a small context, while DROP tasks require answering questions with complex arithmetic computations from the context.\footnote{We compare with baseline performances (i.e., CoT, CoT-SC) reported in previous works if they are higher than our reproduced ones.} Although reasoning in both tasks is largely independent, LLMs still need to align their reasoning with the context. Our method, presented in Table \ref{tab:result-logiqa} for LogiQA and Table \ref{tab:result-drop} for DROP, robustly enhances real-time contextual reasoning in both benchmarks, achieving new state-of-the-art 0-shot results. In both benchmarks, \CoElong~  significantly outperformed existing approaches. 

\begin{table}[!ht]
\centering
\resizebox{\linewidth}{!}{
    \begin{tabular}{l|c|c|c}
        \toprule
         Backbone & Method & Acc & Steps \\  
        \midrule
      \multirow{5}{*}{GPT-4}  &  CoT$^a$ & 38.6  & 1 \\
        & CoT-SC$^a$ & 40.4 & 16 \\
       &  ToT$^a$ & 43.0 & 19.87\\
      &  CR$^a$ & 45.3 & 17 \\
       & \CoElong~  & {\textbf{53.8\%}} & {1}\\ 
        \hline 
        \multirow{3}{*}{PaLM-2}  & CoT & 35.0 & 1 \\
         & CoT-SC  & 36.3 & 3 \\
        & \CoElong~  & 36.4  & 1 \\
        \hline
        \textsc{Previous SOTA$^b$} & - & 45.8 & -\\

        
        \bottomrule
    \end{tabular}
}
\vspace{-10pt}
\caption{Performance on LogiQA. $^{a-b}$ refer to \citet{zhang2023cumulative} and \citet{logiqa_sota} respectively.  }
\label{tab:result-logiqa}
\end{table}

\newcommand{\hlc}[2]{\colorbox{#1}{#2}}

\begin{table*}[ht]
\small 
\centering
\begin{tabular}{p{4cm}p{11cm}}
\toprule
\textbf{Q:} & How many field goals of at least 40 yards did Neil Rackers make \\
\textbf{Ans:} & [`2', `2'] \\
\midrule
\textbf{Context:} & .. In the fourth quarter, the Seahawks took the lead with Brown 
kicking another 28-yard field goal. However, Arizona rallied to tie the 
game on a \hlc{contextHighlight}{Rackers 52-yard}.. Arizona quickly moved into field goal range 
after the Seahawk TD, but \hlc{contextHighlight}{Rackers hit the upright from 53 yards} as time 
expired. .. Arizona then drove to the Seahawks' 25-yard line, where 
\hlc{contextHighlight}{Rackers kicked a game-winning 42-yard field goal} with 11 seconds left. \\
\midrule
\textbf{CoT Reasonings:} & The question is asking for the number of field goals of at 
least 40 yards that Neil Rackers made. The context states that Rackers 
made a \hlc{cotHighlight}{52 yard field goal}, a \hlc{cotHighlight}{42-yard field goal}, and a \hlc{cotHighlight}{53-yard field goal}. 
Therefore, the answer is 3. \\
\midrule
\textbf{CoE-Short Reasonings:} & The question is asking for the number of field 
goals of at least 40 yards that Neil Rackers made. The context states that 
Rackers made a \hlc{coeHighlight}{52 yard field goal} and a \hlc{coeHighlight}{42-yard field goal}. Hence, the 
answer is 2. \\
\bottomrule
\end{tabular}
\caption{CoT distracted by focusing on numerical precision only. \CoElong~  provides superior reasoning by considering both arithmetic  and validity of rationales.}
\label{fig:drop-eg2-sup-cot-bad}
\end{table*}

For instance, in \Cref{tab:result-logiqa} using GPT-4 as backbone \CoElong~   achieves 9\% and 11\% higher Acc than CR and ToT respectively on LogiQA while their iterations are much higher in number.  This reveals that variants built on CoT also suffer from generating  outputs inconsistent to context, and guiding their reasoning paths with grounding precision can enhance CoT approaches broadly. We find that while CoT prompts give decisions for MCQ options directly in every step, \CoElong~ explains how the option can/not be inferred from the context (example: Appendix Fig \ref{fig:appendix:example-2:e2g-cot-logiqa}).
Similarly,  \Cref{fig:drop-eg2-sup-cot-bad} shows an example how \toolnew provides superior reasoning w.r.t CoT (more in Appendix). On DROP, PaLM-2 achieves higher performances than GPT-4 in general, and with \CoElong~  it outperforms the few-shot F1 scores of recent performer LLM  Gemini Ultra.

Besides, in compare to the best performances of \CoElong~ in these two tasks, F1 performances of \CoEshort~are (LogiQA 53.8 vs 51.8) and (83.3 vs 82.7) which validates our intuition that \CoElong~  excels more when the task is based on arithmetic and logical reasoning. 
In addition, replacing the \CoElong~  with \CoEshort,  we observe a performance drop of around 2\% \& 0.6\% in  LogiQA amd DROP respectively-- which validates our intuition that \CoElong~  reasoning is both more context-driven and modular combining both the \CoEshort~ and CoT. In simple math tasks (e.g., GSM8K), our method performs as good as CoT as they are  often context-free.



\begin{table}[!ht]
\vspace{-1mm}
\centering
\resizebox{\linewidth}{!}{
    \begin{tabular}{l|c|c|c}
        \toprule
         Backbone & Method & EM & F1 \\  
        \midrule
      \multirow{2}{*}{GPT-4}  &  CoT & 56.2 & 71.3 \\
       & \CoElong~  & 56.4 & {73.7}\\
        \hline 
        \multirow{2}{*}{PaLM-2}  & CoT & - & 82.0$^{a}$\\
        & \CoElong~  & \textbf{79.6} & \textbf{83.3} \\
        \hline
        \textsc{few-shot SOTA} &  & - & 82.4$^{a}$/83.0$^{b}$\\
        
        \bottomrule
    \end{tabular}
}
\caption{Results on DROP. $^{a-b}$ refer to Gemini Technical Report \cite{team2023gemini} and \citet{huang2022large}. }
\label{tab:result-drop}
\end{table}

\begin{table}[ht]
\vspace{-1mm}
\small 
\centering
\resizebox{0.5\textwidth}{!}{
    \begin{tabular}{l|c|c|c|c}
        \toprule
          \multirow{2}{*}{Backbone} &  \multirow{2}{*}{Method} & \multicolumn{2}{c|}{HotpotQA} &  FEVER \\ 
          & & EM & F1 & {Acc}\\  
        \midrule
      \multirow{3}{*}{ChatGPT}  &  CoT & 43.4 & 55.3 & 76.7 \\
       & Rct+Rfl$^3$ (t=2) & 42 & - & -\\
      & \tool & \textbf{47.1} & \textbf{59.6} & 80.7\\
        \hline 


        \multirow{2}{*}{PaLM-2}& CoT &	44.49 &	55.76	& 78.0 \\
 &  \toolnospace & {	46.76}&	{57.90}	& \textbf{82.0} \\

        \hline
        \textsc{SOTA} & - & 72.7$^1$	& 85.0$^1$ & 94.2 $^2$\\

        
        \bottomrule
    \end{tabular}
}
\caption{Performance on HotpotQA. $^{1-3}$ refers to
\citet{hotpotqa_sota_zhang2023beam}, \citet{filco}, \citet{shinn2023reflexion} respectively. \texttt{t=2} refers to our equivalent  2 trials. Rct, Rfl denote  ReACT, and Reflexion respectively. 
}
\vspace{-1mm}
\label{tab:result-hotpotqa-fever}
\end{table}

\minisection{Multi-hop QA with Distracting Contexts}
\label{sec:exp:multi-hopqa}
We tackle more complex QA challenges, evaluating on the distractor split of HotpotQA \citep{yang-etal-2018-hotpotqa}, where each query faces a large context with two relevant and eight irrelevant documents, with only 2-5 far-apart sentences serving as rationales. 
Results in Table \ref{tab:result-hotpotqa-fever} show that \tool, using both ChatGPT and PaLM-2, outperforms CoT and other variants by a large margin. 
With ChatGPT, \tool achieves a 4\% gain in both EM and F1 scores over CoT, while with PaLM-2, the gains are 2\%.
Other iterative baselines like ReACT and Reflexion find the relevant texts one by one, leading to longer trials or failure to utilize all relevant hops effectively.
Note that with PaLM-2, their  literature reported performances are found with PaLM-1 and we do not reproduce them as they use few-shot examples, leverage external tools, and involve iterative generations. In contrast, both E and G steps of \tool address the entire problem in each step, demonstrating greater efficiency.  As connecting the relevant texts distant apart is one of the key challenges in multi-hop QA, to understand the advantages of extracting \texttt{Evidence} along with the answer (i.e., single objective in both steps), we perform an additional experiment towards a multi-agent or multi-objective approach. 
We employ ChatGPT to establish one agent tasked with co-reference resolution, another as a retriever model for sentence ranking, and a third as a QA agent for generating the final answer. However, this approach lags behind by 10EM points compared to ours. We identify multiple issues, including each agent contributing to propagating its own errors such as imperfect co-reference resolution and ranking. We also find that the agents are overly sensitive to input objective instructions.


\begin{table*}[ht]
\small 
\centering
\begin{tabular}{p{4cm}p{11cm}}
\toprule
\textbf{Q:} & Are both Dafeng District and Dazhou located in the same province? \\
\textbf{Ans:} & \hlc{contextHighlight}{No} \\
\midrule
\textbf{CoT Reasonings:} &Dafeng District and Dazhou are both located in the same
province, which is Sichuan province. \hlc{cotHighlight}{False information is given in the
context} that {Dafeng and Dazhou are located in different provinces.}  \\
\textbf{Ans:} & \hlc{cotHighlight}{Yes} \\
\midrule
\textbf{\CoEshort~ Reasonings:} & {Dafeng District and Dazhou are not located in the
same province.} \hlc{coeHighlight}{Dafeng District is in Jiangsu province} while \hlc{coeHighlight}{Dazhou is in Sichuan province.} \\
\textbf{Ans:} & \hlc{coeHighlight}{No} \\
\bottomrule
\end{tabular}
\caption{CoT reasoning does not account for grounding--even overrides the  contextual fact with its pre-trained knowledge (or hallucinates). Finding \texttt{Evidence}, \tool enables more factually verified reasoning.}
\label{fig:hotpotqa-example-results}
\end{table*}

\begin{table*}[ht]
\vspace{-1mm}
\small 
\centering
    \begin{tabular}{l|c|c|c|c|c|c|c}
        \toprule
          \multirow{2}{*}{Backbone} &  \multirow{2}{*}{Method} & \multicolumn{2}{c|}{NQ} & \multicolumn{2}{c|}{TQA}  & {WOW} & ELI5  \\ 
          & & EM & F1 & EM & F1  & F1 &  F1 \\  
        \midrule
      \multirow{2}{*}{ChatGPT} &  CoT & 41.6 & 51.9 & 68.3 & 75.4 & 13.4 & \textbf{27.0} \\
      & \toolnospace & \textbf{42.8} & \textbf{53.0} &\textbf{69.5} & \textbf{76.9} & \textbf{15.0} & {\color{red}{25.1}}\\
        \hline 
        \multirow{2}{*}{PaLM-2}  & CoT  & 28.4 & 36.6. & 46.9 & 51.9  & 12.2 & 15.3\\ 
        &  \toolnospace & 31.2 & 39.5 & 46.7 & 52.1 & 12.4  & 17.4 \\
        \hline
        \textsc{Sup. SOTA$^1$} & & & 61.8	& - & 71.1 &  68.3  & 73.9 \\
    
        \bottomrule
    \end{tabular}
\caption{Results on NQ, TQA, WOW, and ELI5. $^1$ \& {\color{red}{Red}} refer to \citet{filco} \&  an  inferior performance.}
\vspace{-1mm}
\label{tab:result-nq-tqa-eli5_wow}
\end{table*}


\begin{table*}[ht]
\small 
\centering
\begin{tabular}{p{4.5cm}p{11cm}}
\toprule
\textbf{Q:} &  Who was in dont worry be happy video? \\
\textbf{Ans:} & \hlc{contextHighlight}{['Bill Irwin', 'Robin Williams', 'McFerrin']} \\
\midrule
\textbf{E-Step (CoE-Short) Reasonings:}& The comedic original video for
'Don't Worry Be Happy' stars Bobby McFerrin, Robin Williams, and
Bill Irwin.   \\
\textbf{Ans:} & \hlc{cotHighlight}{Robin Williams} \\
\midrule
\textbf{G-Step (CoE-Short) Reasonings} & The video for 'Don't Worry
Be Happy' stars Robin Williams and Bill Irwin along with
McFerrin. \\
\textbf{Ans:} & \hlc{coeHighlight}{Robin Williams and Bill Irwin} \\
\bottomrule
\end{tabular}
\caption{E-step may focus on answering partially when
asked joint questions or multiple named entity answers.
Hence, to increase our chances, in second step (G) Context, we use the \texttt{Evidence + Original Context}.}
\label{fig:nq-problem}
\end{table*}

In addition, a key bottleneck arises from the retriever agent as it is unaware of how its outputs will be combined by the later QA agent, leading to sub-optimal ranking. For instance, when queried about two persons, all top-$k$ sentences pertaining to one same person may be ranked higher than those about the other, adding complexity to the task.  In addition, we observe some interesting hallucination trends with CoT when the context contains distractions:  LLMs' hallucination even override the factual information in the context. \Cref{fig:hotpotqa-example-results} illustrates this with an example where  \tool constructively emphasizes on evidences and tackles this. 
In a further experiment, we find an increase of 5 points both EM and F1 score when using \CoElong~ instead of \CoEshort~--validating its higher effectiveness.

\minisection{Retrieval Augmented Generation}
\label{sec:exp:rag}
In addition to the MRC and Distractor, we evaluate our framework on the following five RAG tasks in the KILT benchmark \citep{petroni-etal-2021-kilt}.

\noindent{\bf Fact Verification:} 
We adopt the Fact Extraction and VERification (FEVER) dataset \cite{thorne-etal-2018-fever}. The task involves determining whether a claim aligns with facts in a Wikipedia reference ("SUPPORTS") or contradicts them ("REFUTES"). As shown in Table \ref{tab:result-hotpotqa-fever}, \tool outperforms strong baselines by more than 4\% across both LLMs. Further comparisons with CoT-SC (Self-consistency; \citep{wang2022self}) validate that performance gaps of over 2\% persist. Our \texttt{Evidence} captures essential rationales for claim evaluation, and akin to HotpotQA, our global problem-solving approach provides advantages over iterative CoT variants (FEVER reasoning examples  are in Appendix).

\noindent{\bf Open-Domain Question Answering:} 
We adopt the Natural Questions (NQ) \cite{kwiatkowski-etal-2019-natural} and TriviaQA (TQA) \cite{joshi-etal-2017-triviaqa} benchmark to analyze our prompting framework. For each example, there is a short associated answers (less than five tokens) to generate. We present model performances w/ \tool in Table \ref{tab:result-nq-tqa-eli5_wow}. 
We note that questions in NQ are often joint or has multiple named entity answers, and hence we choose to the analyze the affect of different alternatives for the G-Step context.  As shown in \Cref{fig:nq-problem}, LLMs outputs can answer partially in E-step and using \texttt{Evidence + Original Context} as G-step context provides additional chances to the model for answering the query fully - consequently enhances  model enhances. 
\begin{figure}[ht]
    \centering
    \includegraphics[width=0.4\textwidth]{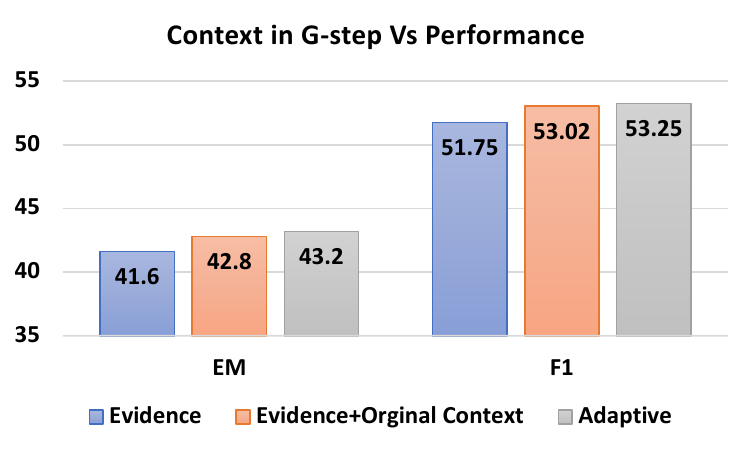}
    \caption{Reasoning with different  "Context" in G-step. Adaptive means selecting them dynamically on the fly.}
\label{fig:nq-adaptive}
\end{figure}
To grasp more on improving \toolnospace, investigating the dataset more, we further develop an adaptive G-step context with a heuristic that the well formed questions (starts with \emph{wh} words or "be/do/have/modal" words) are often from this multi-query type. 
Hence using \texttt{Evidence + Original Context} for them and using only \texttt{Evidence)} for  ill-formed others (e.g., my age is what?)--leading to even better performances.  Figure \ref{fig:nq-adaptive} compares them.

\begin{table*}[ht]
\small 
\centering
\begin{tabular}{p{4cm}p{11cm}}
\toprule
\textbf{Q:} & Sounds complicated. How long did it take you to learn to do that?
\\
\textbf{Ans:} & \hlc{contextHighlight}{
It depends on how fast you can learn the languages such as COBOL, C,
C++, C\#, } \hlc{contextHighlight}{ Java, Lisp, Python, etc
} 
\\
\midrule
\textbf{CoT Reasonings:} & The context provided discusses various topics such as
game programming..The user's claim does not relate to any of these topics.
\\
\textbf{Ans:} & \hlc{cotHighlight}{I am sorry, but your claim does not relate to the context provided.} \\
\midrule
\textbf{Our final G-Step Reasonings:} &  Based on the context, it is mentioned that
software development takes weeks or months to complete, but it depends
on the complexity of the software. Some complex software may require
more than a year of work, while others are..
\\
\textbf{Ans:} & \hlc{coeHighlight}{Learning to create
software can take varying amounts of time depending on the  }
\hlc{coeHighlight}{ complexity of the software and the individual's learning ability.} \\
\bottomrule
\end{tabular}
\caption{Dialogue by ChatGPT using CoT and \toolnospace.}
\label{fig:wow-example}
\end{table*}

\begin{table*}[ht]
\small 
\centering
\begin{tabular}{p{4cm}p{11cm}}
\toprule
\textbf{Q:} & Who has the most trophies in the premier league?
\\
\textbf{Ans:} & \hlc{contextHighlight}{Manchester United} 
\\
\midrule
\textbf{CoT Reasonings:} & Manchester United has won more trophies than any
other club in English football, with a record 20 League titles, 12 FA Cups,
5 League Cups and a ...
\\
\textbf{Ans:} & \hlc{coeHighlight}{Manchester United} \\
\midrule
\textbf{Our final G-Step Reasonings:} & Former Blackburn Rovers and Newcastle
United striker Alan Shearer holds the record for most Premier League goals
with 260...
\\
\textbf{Ans:} & \hlc{cotHighlight}{Alan Sheare} \\
\bottomrule
\end{tabular}
\caption{Overemphasizing on grounding can hinder model from leveraging world knowledge, common sense, etc., (e.g., \hlc{coeHighlight}{Manchester United is a team in premier league}).}
\label{fig:neg-gen-example}
\end{table*}

\noindent{\bf Open-ended Long Form Generation:}
\label{sec:result-wow-example}
Though, our focus is toward the reasoning tasks,  we also explore its potential in open-ended generation tasks. We examine two verbose QA tasks: (i) Knowledge-Grounded Dialog Generation using the WoW dataset \citep{dinan2018wizard}, where short dialog histories are provided as context for generating next-turn responses; (ii) Long Form QA on the ELI5 dataset \citep{fan-etal-2019-eli5}, requiring detailed answers to open-ended questions. Table \ref{tab:result-nq-tqa-eli5_wow} compares \tool with CoT baseline. Although the performance difference is marginal due to verbosity, \tool shows a slight gain over CoT in WoW. A small-scale human evaluation also favored \tool responses by 71\% for factual correctness, similarity to gold responses, and naturalness.  \Cref{fig:wow-example} illustrates a dialogue example, where the overall responses are similar despite verbosity. However, our results in both benchmarks lag behind recent Supervised SOTA models. In Appendix \ref{sec:Appendix:generalizability}, we perform additional experiments and further discuss the generalizability.



\section{Qualitative Case Study: Why and How Our Methods Work? }
\label{sec:cot-error}
To understand more on why and how \toolnew and \tool enhance CoT like reasoning in RAG or with long context, we conduct a case study on CoT reasoning on complex multihop HotpotQA with a set of 50 examples. We observe 4 types of errors: (a) when the question is very hard in reasoning (even for human) (b) when relevant text lies in the middle or at bottom of retrieved context, as noted in \citep{LOST_MIDDLE}. (c)  linguistically or logically challenging questions with long contexts (d) reasoning is not mentioned in the context.  We focus on c, and d.   For problem c,  among the erroneous \emph{wh} questions, in 23\% of them, the gold answer span is actually present in the reasoning, and for the erroneous \emph{yes/no} questions, 75\% of their reasoning actually hypotheses opposite of the predicted answer (e.g., "yes" should be derived from reasoning but the predicted answer is "no"). This indicates that just using the reasoning to answer the question can achieve quite some improvements--justifying our intuition for two-step \tool prompt. For problem d, in our analyses,  23\% of erroneous  \emph{wh} and 25\% of \emph{yes/no} questions are of this category. This suggests a root change in the prompting strategy to focus on verification of the reasoning rationales and to verify, \toolnew shows  an 8\% lower error rate. In addition to qualitative analysis, we employ the self-reflection approach \citep{shinn2023reflexion} by deliberately querying two state-of-the-art LLMs (ChatGPT and Gemini Pro) about the internal advantages of our designed instruction over CoT. Details are provided in Appendix \ref{sec:reflexion}.

\section{Error Analysis and Challenges}
\label{sec: error-analyis}
\begin{figure}[ht]
    \centering
    \includegraphics[width=0.33\textwidth]{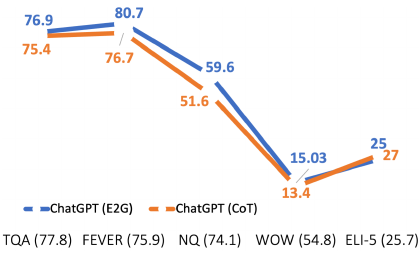}
    \caption{
    \vspace{-4mm}
    F1 scores w/ \tool \& CoT vs (sorted) recall. }
\label{fig:recall-vs-score}
\end{figure}
  Apart from persisted hallucination to some extent, our experiments and ablations reveal two main limitations of our framework. \textbf{Overemphasis in context-grounding} Some overemphasis on grounding leading to the model's failure to infer simple common sense, leverage generic world knowledge, arithmetic, logic, and principles (See \Cref{fig:neg-gen-example}), and in many cases, it causing the model to generate responses such as "unknown," or "cannot be determined". Specific examples of categorical mistakes are provided in the Appendix. \textbf{Low performance in long form generation} We find that the retrieval recalls in WoW and ELI5 are lower than our other RAG tasks (See Figure \ref{fig:recall-vs-score}) which may cause this. Upon investigating more on a performance drop in ELI5: while the task is to generate verbose answers, ours are still short (Word length 130 vs <100) and may actually not fulfilling the target requirements--suggesting a future work of model fine-tuning/domain adaptation.  

\section{Conclusion }
\label{sec:conclusion}
In this paper, we address the limitations of existing prompting frameworks for context-aware and retrieval augmented reasoning. We highlight the challenge of ungrounded reasoning rationales leading to potential hallucinations in LLMs. Our novel framework introduces two new prompting methods to identify evidences in the context and generate answers based on that evidence. Across various tasks, our approach empowers LLMs to deliver robust, and accurate. Future work involves LLM instruction fine-tuning using our prompted outputs. 



\section{Limitations}
Our proposed inference framework has achieved significant gains over baseline approaches across various tasks, and in English. However, in certain data domains (e.g., bio-medical domain \citep{nentidis2023bioasq}), or language (e.g., low-resource languages \cite{parvez2021evaluating}), under automatic evaluation metrics, and with sufficient computational resources or LLMs, it may not exhibit such trends. Another aspect is that the performance scale in RAG tasks may also vary if the retrieval accuracy is quite different than ours. Our evaluation considers the EM, F1, Accuracy, and such matrices for method comparisons, and a different comparison outcomes may be found while using different sets of matrices. For RAG tasks, we use top-5 retrieved documents with any context filtering (e.g., \cite{parvez-etal-2023-retrieval}) and for all tasks, we did  not adopt any model fine-tuning. Under these change in settings, a different kind of results may be obtained regarding which we do not conduct any experiments on. We also note an additional risk of getting different performances on a different number of test instances in the benchmark datasets we reported.





%

\section*{Ethics}
In this paper, we conduct a small scale human evaluation. All our participants were pre-informed about the voluntary nature of our survey, approximated required time, criteria of the feedback. An example human evaluation screen-shot can be found: \url{https://forms.gle/h6WJtC7TrDj9LUNc6}. The participants span different continents, and asked through author's research channels. 

\bibliography{anthology, custom}

\begin{thebibliography}{63}
\expandafter\ifx\csname natexlab\endcsname\relax\def\natexlab#1{#1}\fi

\bibitem[{Asai et~al.(2023{\natexlab{a}})Asai, Min, Zhong, and Chen}]{asai2023retrieval}
Akari Asai, Sewon Min, Zexuan Zhong, and Danqi Chen. 2023{\natexlab{a}}.
\newblock Retrieval-based language models and applications.
\newblock In \emph{Proceedings of the 61st Annual Meeting of the Association for Computational Linguistics (Volume 6: Tutorial Abstracts)}, pages 41--46.

\bibitem[{Asai et~al.(2023{\natexlab{b}})Asai, Wu, Wang, Sil, and Hajishirzi}]{asai2023self}
Akari Asai, Zeqiu Wu, Yizhong Wang, Avirup Sil, and Hannaneh Hajishirzi. 2023{\natexlab{b}}.
\newblock Self-rag: Learning to retrieve, generate, and critique through self-reflection.
\newblock \emph{arXiv preprint arXiv:2310.11511}.

\bibitem[{Creswell et~al.(2022)Creswell, Shanahan, and Higgins}]{creswell2022selection}
Antonia Creswell, Murray Shanahan, and Irina Higgins. 2022.
\newblock Selection-inference: Exploiting large language models for interpretable logical reasoning.
\newblock \emph{arXiv preprint arXiv:2205.09712}.

\bibitem[{Dalvi et~al.(2024)Dalvi, Hasanain, Boughorbel, Mousi, Abdaljalil, Nazar, Abdelali, Chowdhury, Mubarak, Ali, Hawasly, Durrani, and Alam}]{dalvi2023llmebench}
Fahim Dalvi, Maram Hasanain, Sabri Boughorbel, Basel Mousi, Samir Abdaljalil, Nizi Nazar, Ahmed Abdelali, Shammur~Absar Chowdhury, Hamdy Mubarak, Ahmed Ali, Majd Hawasly, Nadir Durrani, and Firoj Alam. 2024.
\newblock {LLMeBench}: A flexible framework for accelerating llms benchmarking.

\bibitem[{Dhuliawala et~al.(2023)Dhuliawala, Komeili, Xu, Raileanu, Li, Celikyilmaz, and Weston}]{dhuliawala2023chain}
Shehzaad Dhuliawala, Mojtaba Komeili, Jing Xu, Roberta Raileanu, Xian Li, Asli Celikyilmaz, and Jason Weston. 2023.
\newblock Chain-of-verification reduces hallucination in large language models.
\newblock \emph{arXiv preprint arXiv:2309.11495}.

\bibitem[{Dinan et~al.(2019)Dinan, Roller, Shuster, Fan, Auli, and Weston}]{dinan2018wizard}
Emily Dinan, Stephen Roller, Kurt Shuster, Angela Fan, Michael Auli, and Jason Weston. 2019.
\newblock \href {https://openreview.net/forum?id=r1l73iRqKm} {{W}izard of {W}ikipedia: Knowledge-powered conversational agents}.
\newblock In \emph{International Conference on Learning Representations}.

\bibitem[{Fan et~al.(2019)Fan, Jernite, Perez, Grangier, Weston, and Auli}]{fan-etal-2019-eli5}
Angela Fan, Yacine Jernite, Ethan Perez, David Grangier, Jason Weston, and Michael Auli. 2019.
\newblock \href {https://doi.org/10.18653/v1/P19-1346} {{ELI}5: Long form question answering}.
\newblock In \emph{Proceedings of the 57th Annual Meeting of the Association for Computational Linguistics}, pages 3558--3567, Florence, Italy. Association for Computational Linguistics.

\bibitem[{Fu et~al.(2022)Fu, Peng, Sabharwal, Clark, and Khot}]{fu2022complexity}
Yao Fu, Hao Peng, Ashish Sabharwal, Peter Clark, and Tushar Khot. 2022.
\newblock Complexity-based prompting for multi-step reasoning.
\newblock In \emph{The Eleventh International Conference on Learning Representations}.

\bibitem[{Gao et~al.(2023)Gao, Madaan, Zhou, Alon, Liu, Yang, Callan, and Neubig}]{gao2023pal}
Luyu Gao, Aman Madaan, Shuyan Zhou, Uri Alon, Pengfei Liu, Yiming Yang, Jamie Callan, and Graham Neubig. 2023.
\newblock Pal: Program-aided language models.
\newblock In \emph{International Conference on Machine Learning}, pages 10764--10799. PMLR.

\bibitem[{He et~al.(2023)He, Liang, Jiao, Zhang, Yang, Wang, Tu, Shi, and Wang}]{he2023exploring}
Zhiwei He, Tian Liang, Wenxiang Jiao, Zhuosheng Zhang, Yujiu Yang, Rui Wang, Zhaopeng Tu, Shuming Shi, and Xing Wang. 2023.
\newblock Exploring human-like translation strategy with large language models.
\newblock \emph{arXiv preprint arXiv:2305.04118}.

\bibitem[{Huang et~al.(2022)Huang, Gu, Hou, Wu, Wang, Yu, and Han}]{huang2022large}
Jiaxin Huang, Shixiang~Shane Gu, Le~Hou, Yuexin Wu, Xuezhi Wang, Hongkun Yu, and Jiawei Han. 2022.
\newblock Large language models can self-improve.
\newblock \emph{arXiv preprint arXiv:2210.11610}.

\bibitem[{Islam et~al.(2024{\natexlab{a}})Islam, Ali, and Parvez}]{islam-etal-2024-mapcoder}
Md.~Ashraful Islam, Mohammed~Eunus Ali, and Md~Rizwan Parvez. 2024{\natexlab{a}}.
\newblock \href {https://doi.org/10.18653/v1/2024.acl-long.269} {{M}ap{C}oder: Multi-agent code generation for competitive problem solving}.
\newblock In \emph{Proceedings of the 62nd Annual Meeting of the Association for Computational Linguistics (Volume 1: Long Papers)}, pages 4912--4944, Bangkok, Thailand. Association for Computational Linguistics.

\bibitem[{Islam et~al.(2025)Islam, Ali, and Parvez}]{islam2025codesim}
Md~Ashraful Islam, Mohammed~Eunus Ali, and Md~Rizwan Parvez. 2025.
\newblock Codesim: Multi-agent code generation and problem solving through simulation-driven planning and debugging.
\newblock \emph{arXiv preprint arXiv:2502.05664}.

\bibitem[{Islam et~al.(2024{\natexlab{b}})Islam, Rahman, Hossain, Hoque, Joty, and Parvez}]{islam-etal-2024-open}
Shayekh~Bin Islam, Md~Asib Rahman, K~S M~Tozammel Hossain, Enamul Hoque, Shafiq Joty, and Md~Rizwan Parvez. 2024{\natexlab{b}}.
\newblock \href {https://aclanthology.org/2024.findings-emnlp.831} {Open-{RAG}: Enhanced retrieval augmented reasoning with open-source large language models}.
\newblock In \emph{Findings of the Association for Computational Linguistics: EMNLP 2024}, pages 14231--14244, Miami, Florida, USA. Association for Computational Linguistics.

\bibitem[{Joshi et~al.(2017)Joshi, Choi, Weld, and Zettlemoyer}]{joshi-etal-2017-triviaqa}
Mandar Joshi, Eunsol Choi, Daniel Weld, and Luke Zettlemoyer. 2017.
\newblock \href {https://doi.org/10.18653/v1/P17-1147} {{T}rivia{QA}: A large scale distantly supervised challenge dataset for reading comprehension}.
\newblock In \emph{Proceedings of the 55th Annual Meeting of the Association for Computational Linguistics (Volume 1: Long Papers)}, pages 1601--1611, Vancouver, Canada. Association for Computational Linguistics.

\bibitem[{Jung et~al.(2022)Jung, Qin, Welleck, Brahman, Bhagavatula, Bras, and Choi}]{jung2022maieutic}
Jaehun Jung, Lianhui Qin, Sean Welleck, Faeze Brahman, Chandra Bhagavatula, Ronan~Le Bras, and Yejin Choi. 2022.
\newblock Maieutic prompting: Logically consistent reasoning with recursive explanations.
\newblock \emph{arXiv preprint arXiv:2205.11822}.

\bibitem[{Karpukhin et~al.(2020)Karpukhin, Oguz, Min, Lewis, Wu, Edunov, Chen, and Yih}]{karpukhin-etal-2020-dense}
Vladimir Karpukhin, Barlas Oguz, Sewon Min, Patrick Lewis, Ledell Wu, Sergey Edunov, Danqi Chen, and Wen-tau Yih. 2020.
\newblock \href {https://doi.org/10.18653/v1/2020.emnlp-main.550} {Dense passage retrieval for open-domain question answering}.
\newblock In \emph{Proceedings of the 2020 Conference on Empirical Methods in Natural Language Processing (EMNLP)}, pages 6769--6781, Online. Association for Computational Linguistics.

\bibitem[{Kim et~al.(2023)Kim, Baldi, and McAleer}]{kim2023language}
Geunwoo Kim, Pierre Baldi, and Stephen McAleer. 2023.
\newblock Language models can solve computer tasks.
\newblock \emph{arXiv preprint arXiv:2303.17491}.

\bibitem[{Kim et~al.(2022)Kim, Cho, Kim, Kim, Yoo, and Lee}]{kim2022self}
Hyuhng~Joon Kim, Hyunsoo Cho, Junyeob Kim, Taeuk Kim, Kang~Min Yoo, and Sang-goo Lee. 2022.
\newblock Self-generated in-context learning: Leveraging auto-regressive language models as a demonstration generator.
\newblock \emph{arXiv preprint arXiv:2206.08082}.

\bibitem[{Kojima et~al.(2022)Kojima, Gu, Reid, Matsuo, and Iwasawa}]{kojima2022large}
Takeshi Kojima, Shixiang~Shane Gu, Machel Reid, Yutaka Matsuo, and Yusuke Iwasawa. 2022.
\newblock Large language models are zero-shot reasoners.
\newblock \emph{Advances in neural information processing systems}, 35:22199--22213.

\bibitem[{Kwiatkowski et~al.(2019)Kwiatkowski, Palomaki, Redfield, Collins, Parikh, Alberti, Epstein, Polosukhin, Devlin, Lee, Toutanova, Jones, Kelcey, Chang, Dai, Uszkoreit, Le, and Petrov}]{kwiatkowski-etal-2019-natural}
Tom Kwiatkowski, Jennimaria Palomaki, Olivia Redfield, Michael Collins, Ankur Parikh, Chris Alberti, Danielle Epstein, Illia Polosukhin, Jacob Devlin, Kenton Lee, Kristina Toutanova, Llion Jones, Matthew Kelcey, Ming-Wei Chang, Andrew~M. Dai, Jakob Uszkoreit, Quoc Le, and Slav Petrov. 2019.
\newblock \href {https://doi.org/10.1162/tacl_a_00276} {Natural questions: A benchmark for question answering research}.
\newblock \emph{Transactions of the Association for Computational Linguistics}, 7:452--466.

\bibitem[{Lester et~al.(2021)Lester, Al-Rfou, and Constant}]{lester-etal-2021-power}
Brian Lester, Rami Al-Rfou, and Noah Constant. 2021.
\newblock \href {https://doi.org/10.18653/v1/2021.emnlp-main.243} {The power of scale for parameter-efficient prompt tuning}.
\newblock In \emph{Proceedings of the 2021 Conference on Empirical Methods in Natural Language Processing}, pages 3045--3059, Online and Punta Cana, Dominican Republic. Association for Computational Linguistics.

\bibitem[{Li et~al.(2022)Li, Zhang, and Zhao}]{li2022self}
Junlong Li, Zhuosheng Zhang, and Hai Zhao. 2022.
\newblock Self-prompting large language models for open-domain qa.
\newblock \emph{arXiv preprint arXiv:2212.08635}.

\bibitem[{Li et~al.(2023)Li, Zhao, Chia, Ding, Bing, Joty, and Poria}]{li2023chainofknowledge}
Xingxuan Li, Ruochen Zhao, Yew~Ken Chia, Bosheng Ding, Lidong Bing, Shafiq Joty, and Soujanya Poria. 2023.
\newblock Chain of knowledge: A framework for grounding large language models with structured knowledge bases.
\newblock \emph{arXiv preprint arXiv:2305.13269}.

\bibitem[{Liu et~al.(2023{\natexlab{a}})Liu, Jiang, Zhang, Liu, Zhang, Biswas, and Stone}]{liu2023llm}
Bo~Liu, Yuqian Jiang, Xiaohan Zhang, Qiang Liu, Shiqi Zhang, Joydeep Biswas, and Peter Stone. 2023{\natexlab{a}}.
\newblock Llm+ p: Empowering large language models with optimal planning proficiency.
\newblock \emph{arXiv preprint arXiv:2304.11477}.

\bibitem[{Liu et~al.(2023{\natexlab{b}})Liu, Lin, Hewitt, Paranjape, Bevilacqua, Petroni, and Liang}]{LOST_MIDDLE}
Nelson~F Liu, Kevin Lin, John Hewitt, Ashwin Paranjape, Michele Bevilacqua, Fabio Petroni, and Percy Liang. 2023{\natexlab{b}}.
\newblock Lost in the middle: How language models use long contexts.
\newblock \emph{arXiv preprint arXiv:2307.03172}.

\bibitem[{Lu et~al.(2021)Lu, Welleck, West, Jiang, Kasai, Khashabi, Bras, Qin, Yu, Zellers et~al.}]{lu2021neurologic}
Ximing Lu, Sean Welleck, Peter West, Liwei Jiang, Jungo Kasai, Daniel Khashabi, Ronan~Le Bras, Lianhui Qin, Youngjae Yu, Rowan Zellers, et~al. 2021.
\newblock Neurologic a* esque decoding: Constrained text generation with lookahead heuristics.
\newblock \emph{arXiv preprint arXiv:2112.08726}.

\bibitem[{Madaan et~al.(2023)Madaan, Tandon, Gupta, Hallinan, Gao, Wiegreffe, Alon, Dziri, Prabhumoye, Yang et~al.}]{madaan2023self}
Aman Madaan, Niket Tandon, Prakhar Gupta, Skyler Hallinan, Luyu Gao, Sarah Wiegreffe, Uri Alon, Nouha Dziri, Shrimai Prabhumoye, Yiming Yang, et~al. 2023.
\newblock Self-refine: Iterative refinement with self-feedback.
\newblock \emph{arXiv preprint arXiv:2303.17651}.

\bibitem[{Madaan et~al.(2024)Madaan, Tandon, Gupta, Hallinan, Gao, Wiegreffe, Alon, Dziri, Prabhumoye, Yang et~al.}]{madaan2024self}
Aman Madaan, Niket Tandon, Prakhar Gupta, Skyler Hallinan, Luyu Gao, Sarah Wiegreffe, Uri Alon, Nouha Dziri, Shrimai Prabhumoye, Yiming Yang, et~al. 2024.
\newblock Self-refine: Iterative refinement with self-feedback.
\newblock \emph{Advances in Neural Information Processing Systems}, 36.

\bibitem[{Miao et~al.(2023)Miao, Teh, and Rainforth}]{miao2023selfcheck}
Ning Miao, Yee~Whye Teh, and Tom Rainforth. 2023.
\newblock Selfcheck: Using llms to zero-shot check their own step-by-step reasoning.
\newblock \emph{arXiv preprint arXiv:2308.00436}.

\bibitem[{Nentidis et~al.(2023)Nentidis, Krithara, Paliouras, Farr{\'e}-Maduell, Lima-L{\'o}pez, and Krallinger}]{nentidis2023bioasq}
Anastasios Nentidis, Anastasia Krithara, Georgios Paliouras, Eul{\`a}lia Farr{\'e}-Maduell, Salvador Lima-L{\'o}pez, and Martin Krallinger. 2023.
\newblock \href {https://link.springer.com/chapter/10.1007/978-3-031-28241-6_66} {Bioasq at~clef2023: The eleventh edition of~the~large-scale biomedical semantic indexing and~question answering challenge}.
\newblock In \emph{Advances in Information Retrieval}.

\bibitem[{Nye et~al.(2021)Nye, Andreassen, Gur-Ari, Michalewski, Austin, Bieber, Dohan, Lewkowycz, Bosma, Luan et~al.}]{nye2021show}
Maxwell Nye, Anders~Johan Andreassen, Guy Gur-Ari, Henryk Michalewski, Jacob Austin, David Bieber, David Dohan, Aitor Lewkowycz, Maarten Bosma, David Luan, et~al. 2021.
\newblock Show your work: Scratchpads for intermediate computation with language models.
\newblock \emph{arXiv preprint arXiv:2112.00114}.

\bibitem[{Ouyang et~al.(2022)Ouyang, Wu, Jiang, Almeida, Wainwright, Mishkin, Zhang, Agarwal, Slama, Ray et~al.}]{ouyang2022training}
Long Ouyang, Jeffrey Wu, Xu~Jiang, Diogo Almeida, Carroll Wainwright, Pamela Mishkin, Chong Zhang, Sandhini Agarwal, Katarina Slama, Alex Ray, et~al. 2022.
\newblock Training language models to follow instructions with human feedback.
\newblock \emph{Advances in Neural Information Processing Systems}, 35:27730--27744.

\bibitem[{Ouyang et~al.(2021)Ouyang, Zhang, and Zhao}]{logiqa_sota}
Siru Ouyang, Zhuosheng Zhang, and Hai Zhao. 2021.
\newblock \href {http://arxiv.org/abs/2105.10334} {Fact-driven logical reasoning}.
\newblock \emph{CoRR}, abs/2105.10334.

\bibitem[{Parvez et~al.(2021)Parvez, Ahmad, Chakraborty, Ray, and Chang}]{parvez-etal-2021-retrieval-augmented}
Md~Rizwan Parvez, Wasi Ahmad, Saikat Chakraborty, Baishakhi Ray, and Kai-Wei Chang. 2021.
\newblock \href {https://doi.org/10.18653/v1/2021.findings-emnlp.232} {Retrieval augmented code generation and summarization}.
\newblock In \emph{Findings of the Association for Computational Linguistics: EMNLP 2021}, pages 2719--2734, Punta Cana, Dominican Republic. Association for Computational Linguistics.

\bibitem[{Parvez and Chang(2021)}]{parvez2021evaluating}
Md~Rizwan Parvez and Kai-Wei Chang. 2021.
\newblock Evaluating the values of sources in transfer learning.
\newblock In \emph{Proceedings of the 2021 Conference of the North American Chapter of the Association for Computational Linguistics: Human Language Technologies}, pages 5084--5116.

\bibitem[{Parvez et~al.(2023)Parvez, Chi, Ahmad, Tian, and Chang}]{parvez-etal-2023-retrieval}
Md~Rizwan Parvez, Jianfeng Chi, Wasi~Uddin Ahmad, Yuan Tian, and Kai-Wei Chang. 2023.
\newblock \href {https://doi.org/10.18653/v1/2023.eacl-main.16} {Retrieval enhanced data augmentation for question answering on privacy policies}.
\newblock In \emph{Proceedings of the 17th Conference of the European Chapter of the Association for Computational Linguistics}, pages 201--210, Dubrovnik, Croatia. Association for Computational Linguistics.

\bibitem[{Paul et~al.(2023)Paul, Ismayilzada, Peyrard, Borges, Bosselut, West, and Faltings}]{paul2023refiner}
Debjit Paul, Mete Ismayilzada, Maxime Peyrard, Beatriz Borges, Antoine Bosselut, Robert West, and Boi Faltings. 2023.
\newblock Refiner: Reasoning feedback on intermediate representations.
\newblock \emph{arXiv preprint arXiv:2304.01904}.

\bibitem[{Petroni et~al.(2021)Petroni, Piktus, Fan, Lewis, Yazdani, De~Cao, Thorne, Jernite, Karpukhin, Maillard, Plachouras, Rockt{\"a}schel, and Riedel}]{petroni-etal-2021-kilt}
Fabio Petroni, Aleksandra Piktus, Angela Fan, Patrick Lewis, Majid Yazdani, Nicola De~Cao, James Thorne, Yacine Jernite, Vladimir Karpukhin, Jean Maillard, Vassilis Plachouras, Tim Rockt{\"a}schel, and Sebastian Riedel. 2021.
\newblock \href {https://doi.org/10.18653/v1/2021.naacl-main.200} {{KILT}: a benchmark for knowledge intensive language tasks}.
\newblock In \emph{Proceedings of the 2021 Conference of the North American Chapter of the Association for Computational Linguistics: Human Language Technologies}, pages 2523--2544, Online. Association for Computational Linguistics.

\bibitem[{Pryzant et~al.(2023)Pryzant, Iter, Li, Lee, Zhu, and Zeng}]{pryzant2023automatic}
Reid Pryzant, Dan Iter, Jerry Li, Yin~Tat Lee, Chenguang Zhu, and Michael Zeng. 2023.
\newblock Automatic prompt optimization with" gradient descent" and beam search.
\newblock \emph{arXiv preprint arXiv:2305.03495}.

\bibitem[{Sanh et~al.(2021)Sanh, Webson, Raffel, Bach, Sutawika, Alyafeai, Chaffin, Stiegler, Scao, Raja et~al.}]{sanh2021multitask}
Victor Sanh, Albert Webson, Colin Raffel, Stephen~H Bach, Lintang Sutawika, Zaid Alyafeai, Antoine Chaffin, Arnaud Stiegler, Teven~Le Scao, Arun Raja, et~al. 2021.
\newblock Multitask prompted training enables zero-shot task generalization.
\newblock \emph{arXiv preprint arXiv:2110.08207}.

\bibitem[{Shinn et~al.(2023)Shinn, Labash, and Gopinath}]{shinn2023reflexion}
Noah Shinn, Beck Labash, and Ashwin Gopinath. 2023.
\newblock Reflexion: an autonomous agent with dynamic memory and self-reflection.
\newblock \emph{arXiv preprint arXiv:2303.11366}.

\bibitem[{Sun et~al.(2022)Sun, Wang, Tay, Yang, and Zhou}]{sun2022recitation}
Zhiqing Sun, Xuezhi Wang, Yi~Tay, Yiming Yang, and Denny Zhou. 2022.
\newblock Recitation-augmented language models.
\newblock \emph{arXiv preprint arXiv:2210.01296}.

\bibitem[{Team et~al.(2023)Team, Anil, Borgeaud, Alayrac, Yu, Soricut, Schalkwyk, Dai, Hauth, Millican et~al.}]{team2023gemini}
Gemini Team, Rohan Anil, Sebastian Borgeaud, Jean-Baptiste Alayrac, Jiahui Yu, Radu Soricut, Johan Schalkwyk, Andrew~M Dai, Anja Hauth, Katie Millican, et~al. 2023.
\newblock Gemini: a family of highly capable multimodal models.
\newblock \emph{arXiv preprint arXiv:2312.11805}.

\bibitem[{Thorne et~al.(2018)Thorne, Vlachos, Christodoulopoulos, and Mittal}]{thorne-etal-2018-fever}
James Thorne, Andreas Vlachos, Christos Christodoulopoulos, and Arpit Mittal. 2018.
\newblock \href {https://doi.org/10.18653/v1/N18-1074} {{FEVER}: a large-scale dataset for fact extraction and {VER}ification}.
\newblock In \emph{Proceedings of the 2018 Conference of the North {A}merican Chapter of the Association for Computational Linguistics: Human Language Technologies, Volume 1 (Long Papers)}, pages 809--819, New Orleans, Louisiana. Association for Computational Linguistics.

\bibitem[{Trivedi et~al.(2023)Trivedi, Balasubramanian, Khot, and Sabharwal}]{trivedi-etal-2023-interleaving}
Harsh Trivedi, Niranjan Balasubramanian, Tushar Khot, and Ashish Sabharwal. 2023.
\newblock \href {https://doi.org/10.18653/v1/2023.acl-long.557} {Interleaving retrieval with chain-of-thought reasoning for knowledge-intensive multi-step questions}.
\newblock In \emph{Proceedings of the 61st Annual Meeting of the Association for Computational Linguistics (Volume 1: Long Papers)}, pages 10014--10037, Toronto, Canada. Association for Computational Linguistics.

\bibitem[{Wang et~al.(2023{\natexlab{a}})Wang, Sun, Chen, Li, and Gao}]{wang2023boosting-cok}
Jianing Wang, Qiushi Sun, Nuo Chen, Xiang Li, and Ming Gao. 2023{\natexlab{a}}.
\newblock Boosting language models reasoning with chain-of-knowledge prompting.
\newblock \emph{arXiv preprint arXiv:2306.06427}.

\bibitem[{Wang et~al.(2022)Wang, Wei, Schuurmans, Le, Chi, Narang, Chowdhery, and Zhou}]{wang2022self}
Xuezhi Wang, Jason Wei, Dale Schuurmans, Quoc Le, Ed~Chi, Sharan Narang, Aakanksha Chowdhery, and Denny Zhou. 2022.
\newblock Self-consistency improves chain of thought reasoning in language models.
\newblock \emph{arXiv preprint arXiv:2203.11171}.

\bibitem[{Wang et~al.(2023{\natexlab{b}})Wang, Araki, Jiang, Parvez, and Neubig}]{filco}
Zhiruo Wang, Jun Araki, Zhengbao Jiang, Md~Rizwan Parvez, and Graham Neubig. 2023{\natexlab{b}}.
\newblock Learning to filter context for retrieval-augmented generation.
\newblock \emph{arXiv preprint arXiv:2311.08377}.

\bibitem[{Wei et~al.(2021)Wei, Bosma, Zhao, Guu, Yu, Lester, Du, Dai, and Le}]{wei2021finetuned}
Jason Wei, Maarten Bosma, Vincent~Y Zhao, Kelvin Guu, Adams~Wei Yu, Brian Lester, Nan Du, Andrew~M Dai, and Quoc~V Le. 2021.
\newblock Finetuned language models are zero-shot learners.
\newblock \emph{arXiv preprint arXiv:2109.01652}.

\bibitem[{Wei et~al.(2022)Wei, Wang, Schuurmans, Bosma, Xia, Chi, Le, Zhou et~al.}]{cot}
Jason Wei, Xuezhi Wang, Dale Schuurmans, Maarten Bosma, Fei Xia, Ed~Chi, Quoc~V Le, Denny Zhou, et~al. 2022.
\newblock Chain-of-thought prompting elicits reasoning in large language models.
\newblock \emph{Advances in Neural Information Processing Systems}, 35:24824--24837.

\bibitem[{Xie et~al.(2023)Xie, Kawaguchi, Zhao, Zhao, Kan, He, and Xie}]{xie2023decomposition}
Yuxi Xie, Kenji Kawaguchi, Yiran Zhao, Xu~Zhao, Min-Yen Kan, Junxian He, and Qizhe Xie. 2023.
\newblock Decomposition enhances reasoning via self-evaluation guided decoding.
\newblock \emph{arXiv preprint arXiv:2305.00633}.

\bibitem[{Yang et~al.(2018)Yang, Qi, Zhang, Bengio, Cohen, Salakhutdinov, and Manning}]{yang-etal-2018-hotpotqa}
Zhilin Yang, Peng Qi, Saizheng Zhang, Yoshua Bengio, William Cohen, Ruslan Salakhutdinov, and Christopher~D. Manning. 2018.
\newblock \href {https://doi.org/10.18653/v1/D18-1259} {{H}otpot{QA}: A dataset for diverse, explainable multi-hop question answering}.
\newblock In \emph{Proceedings of the 2018 Conference on Empirical Methods in Natural Language Processing}, pages 2369--2380, Brussels, Belgium. Association for Computational Linguistics.

\bibitem[{Yao et~al.(2023)Yao, Yu, Zhao, Shafran, Griffiths, Cao, and Narasimhan}]{yao2023tree}
Shunyu Yao, Dian Yu, Jeffrey Zhao, Izhak Shafran, Thomas~L Griffiths, Yuan Cao, and Karthik Narasimhan. 2023.
\newblock Tree of thoughts: Deliberate problem solving with large language models.
\newblock \emph{arXiv preprint arXiv:2305.10601}.

\bibitem[{Yao et~al.(2022)Yao, Zhao, Yu, Du, Shafran, Narasimhan, and Cao}]{yao2022react}
Shunyu Yao, Jeffrey Zhao, Dian Yu, Nan Du, Izhak Shafran, Karthik Narasimhan, and Yuan Cao. 2022.
\newblock React: Synergizing reasoning and acting in language models.
\newblock \emph{arXiv preprint arXiv:2210.03629}.

\bibitem[{Yasunaga et~al.(2023)Yasunaga, Chen, Li, Pasupat, Leskovec, Liang, Chi, and Zhou}]{yasunaga2023large}
Michihiro Yasunaga, Xinyun Chen, Yujia Li, Panupong Pasupat, Jure Leskovec, Percy Liang, Ed~H Chi, and Denny Zhou. 2023.
\newblock Large language models as analogical reasoners.
\newblock \emph{arXiv preprint arXiv:2310.01714}.

\bibitem[{Yasunaga et~al.(2024)Yasunaga, Chen, Li, Pasupat, Leskovec, Liang, Chi, and Zhou}]{yasunaga2024large}
Michihiro Yasunaga, Xinyun Chen, Yujia Li, Panupong Pasupat, Jure Leskovec, Percy Liang, Ed~H. Chi, and Denny Zhou. 2024.
\newblock \href {https://openreview.net/forum?id=AgDICX1h50} {Large language models as analogical reasoners}.
\newblock In \emph{The Twelfth International Conference on Learning Representations}.

\bibitem[{Yu et~al.(2020)Yu, Jiang, Dong, and Feng}]{yu2020reclor}
Weihao Yu, Zihang Jiang, Yanfei Dong, and Jiashi Feng. 2020.
\newblock Reclor: A reading comprehension dataset requiring logical reasoning.
\newblock \emph{arXiv preprint arXiv:2002.04326}.

\bibitem[{Yu et~al.(2023)Yu, Zhang, Pan, Ma, Wang, and Yu}]{yu2023chain-of-note}
Wenhao Yu, Hongming Zhang, Xiaoman Pan, Kaixin Ma, Hongwei Wang, and Dong Yu. 2023.
\newblock Chain-of-note: Enhancing robustness in retrieval-augmented language models.
\newblock \emph{arXiv preprint arXiv:2311.09210}.

\bibitem[{Zhang et~al.(2023{\natexlab{a}})Zhang, Zhang, Zhang, Liu, and Huang}]{hotpotqa_sota_zhang2023beam}
Jiahao Zhang, Haiyang Zhang, Dongmei Zhang, Yong Liu, and Shen Huang. 2023{\natexlab{a}}.
\newblock Beam retrieval: General end-to-end retrieval for multi-hop question answering.
\newblock \emph{arXiv preprint arXiv:2308.08973}.

\bibitem[{Zhang et~al.(2023{\natexlab{b}})Zhang, Yang, Yuan, and Yao}]{zhang2023cumulative}
Yifan Zhang, Jingqin Yang, Yang Yuan, and Andrew Chi-Chih Yao. 2023{\natexlab{b}}.
\newblock Cumulative reasoning with large language models.
\newblock \emph{arXiv preprint arXiv:2308.04371}.

\bibitem[{Zhao et~al.(2023)Zhao, Li, Joty, Qin, and Bing}]{zhao2023verify}
Ruochen Zhao, Xingxuan Li, Shafiq Joty, Chengwei Qin, and Lidong Bing. 2023.
\newblock Verify-and-edit: A knowledge-enhanced chain-of-thought framework.
\newblock \emph{arXiv preprint arXiv:2305.03268}.

\bibitem[{Zhu et~al.(2022)Zhu, Wang, Zhang, Zhang, Gan, Zhang, and Yang}]{zhu2022solving}
Xinyu Zhu, Junjie Wang, Lin Zhang, Yuxiang Zhang, Ruyi Gan, Jiaxing Zhang, and Yujiu Yang. 2022.
\newblock Solving math word problem via cooperative reasoning induced language models.
\newblock \emph{arXiv preprint arXiv:2210.16257}.

\end{thebibliography}

\appendix
\section{Appendix}
\label{sec:appendix}


\subsection{Additional Experiments on Generalizability }
\label{sec:Appendix:generalizability}
To further discuss the generalizability of our approach with new LLM models and on new benchmark datasets, we consider two additional experiments on logical/factual reasoning tasks on a randomly sampled 100 examples from (i) LogiQA and DROP dataset using Gemini-Pro as backbone foundation model in Table \ref{tab:gemini-pro} (ii) ReCLOR dataset \cite{yu2020reclor} using ChatGPT as backbone LLM in Table \ref{tab:reclor}. 

\begin{table}[t]
    \centering
    \begin{tabular}{c|c|c|c}
        \multirow{2}{*}{Method} &  	LogiQA & 	\multicolumn{2}{c}{DROP}   \\
        & Acc & EM & F1 \\
         \hline 
CoT	& 35.0	& 51.0 & 	62.06 \\
  \CoElong~ \ & 	\textbf{41.0} &	\textbf{52.0}	& \textbf{63.77} \\
 \hline
    \end{tabular}
    \caption{Results on LogiQA and DROP using Gemini-Pro.}
    \label{tab:gemini-pro}
\end{table}
\begin{table}[t]
    \centering
    \begin{tabular}{c|c}
      Method &  ReCLOR (Acc) \\
         \hline 
CoT	& 46 \\
CoT-SC & 49 \\
  \CoElong~ \ & 	\textbf{53.0} \\
 \hline
    \end{tabular}
    \caption{Results on ReCLOR using ChatGPT.}
    \label{tab:reclor}
\end{table}

\begin{figure}[h]
    \centering
    \includegraphics[width=2\linewidth]{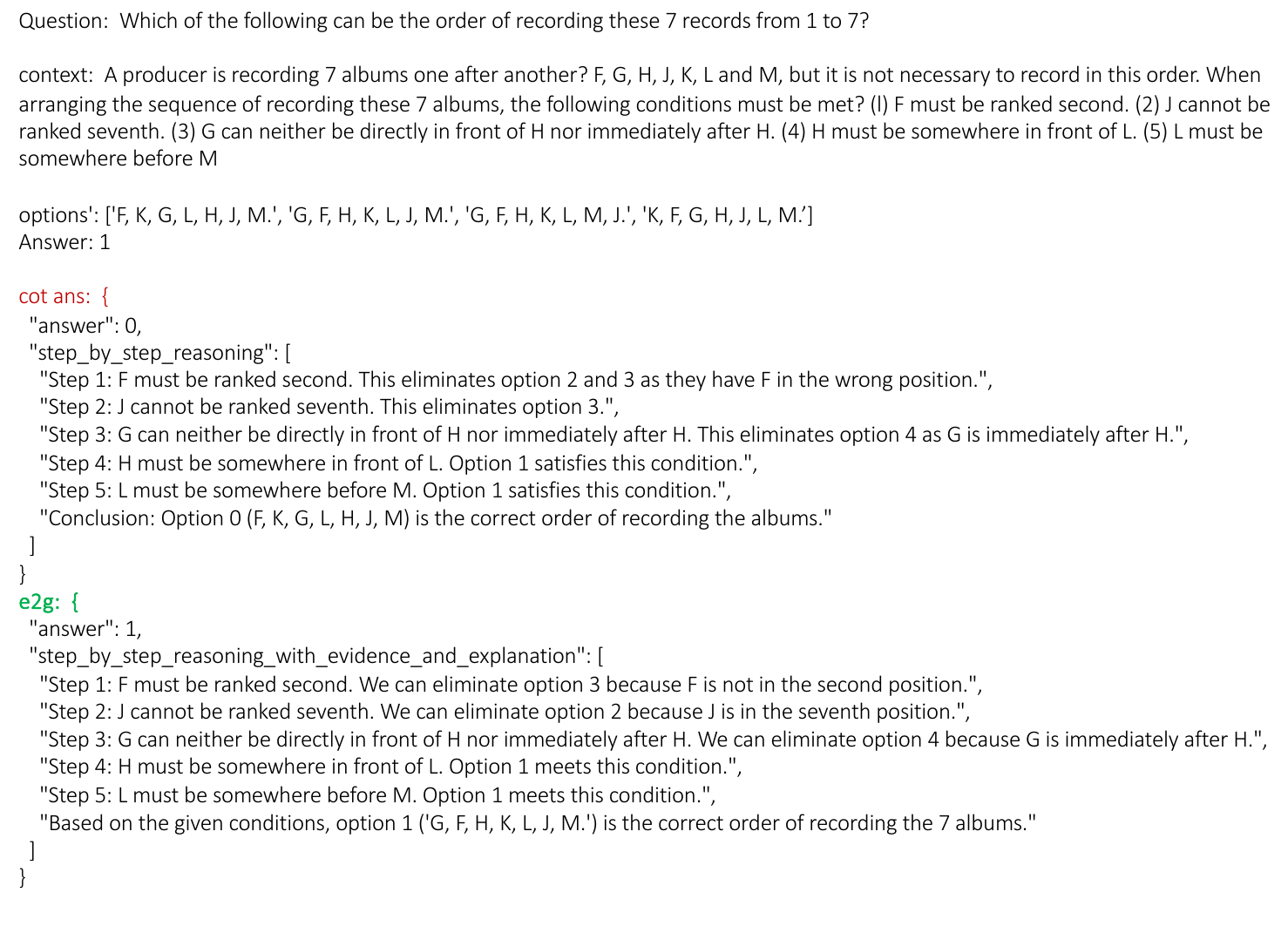}
    \caption{Example-1 w/ different prompting for LogiQA benchmarks}
    \label{fig:appendix:example-2:e2g-cot-logiqa}
\end{figure}

\subsection{Self-Reflection: Why \toolnew Works?}
\label{sec:reflexion}
Along with statistical motivation, to further understand why it works, we consider the self-reflection \cite{shinn2023reflexion} approach--deliberately asking two different SoTA LLMs (ChaTGPT and Gemini Pro) the internal advantages of our designed instruction over CoT. Below we summarize them. 
\begin{enumerate}
    \itemsep0em 
    \small 
     \item {\textbf{Logical Reasoning:} promotes more structured and logical thought process, reducing unsupported 
     statements.}
    \item {\textbf{Factual Basis:} Explicitly asking to focus on justifying its answer by providing evidence \& explanation encourages the LLM to ground its reasoning in the context and relevant facts, making it less likely to resort to imaginary or unsupported claims.}
    \item {\textbf{Reduced Speculation:} Prompting for evidence encourages to rely on what is known or can be reasonably inferred from existing information. }
     \item {\textbf{Accountability:} When prompted to provide evidence, models are held accountable for the accuracy and reliability of their responses. }
\end{enumerate}

\subsection{Qualitative Examples}
\label{sec:qualitative-exmples-appendix}


\begin{figure*}
    \centering
    \includegraphics[width=\linewidth]{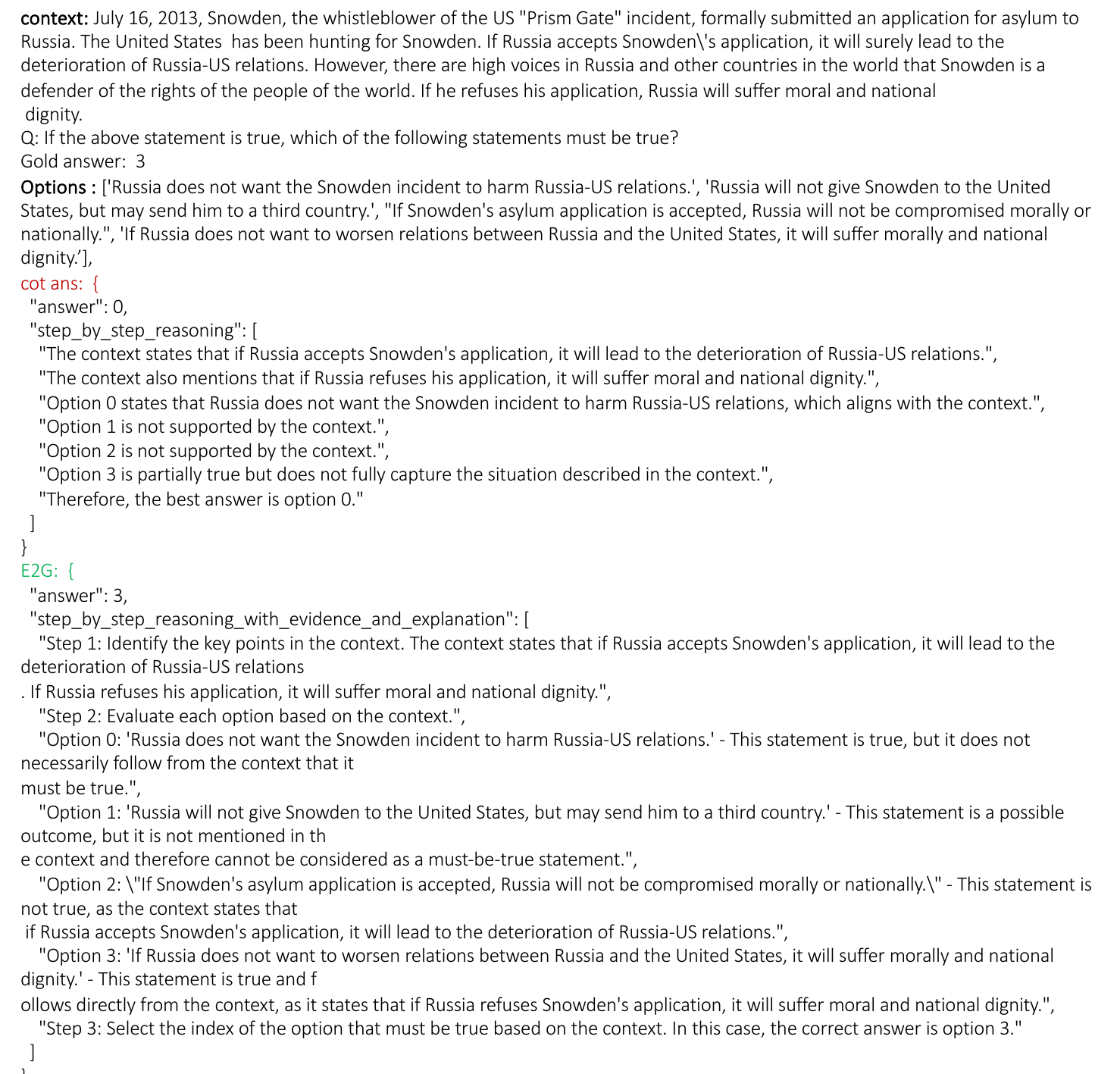}
    \caption{Example-2 w/ different prompting for LogiQA benchmarks}
    \label{fig:appendix:example-1:e2g-cot-logiqa}
\end{figure*}

\begin{figure*}
    \centering
    \includegraphics[width=\linewidth]{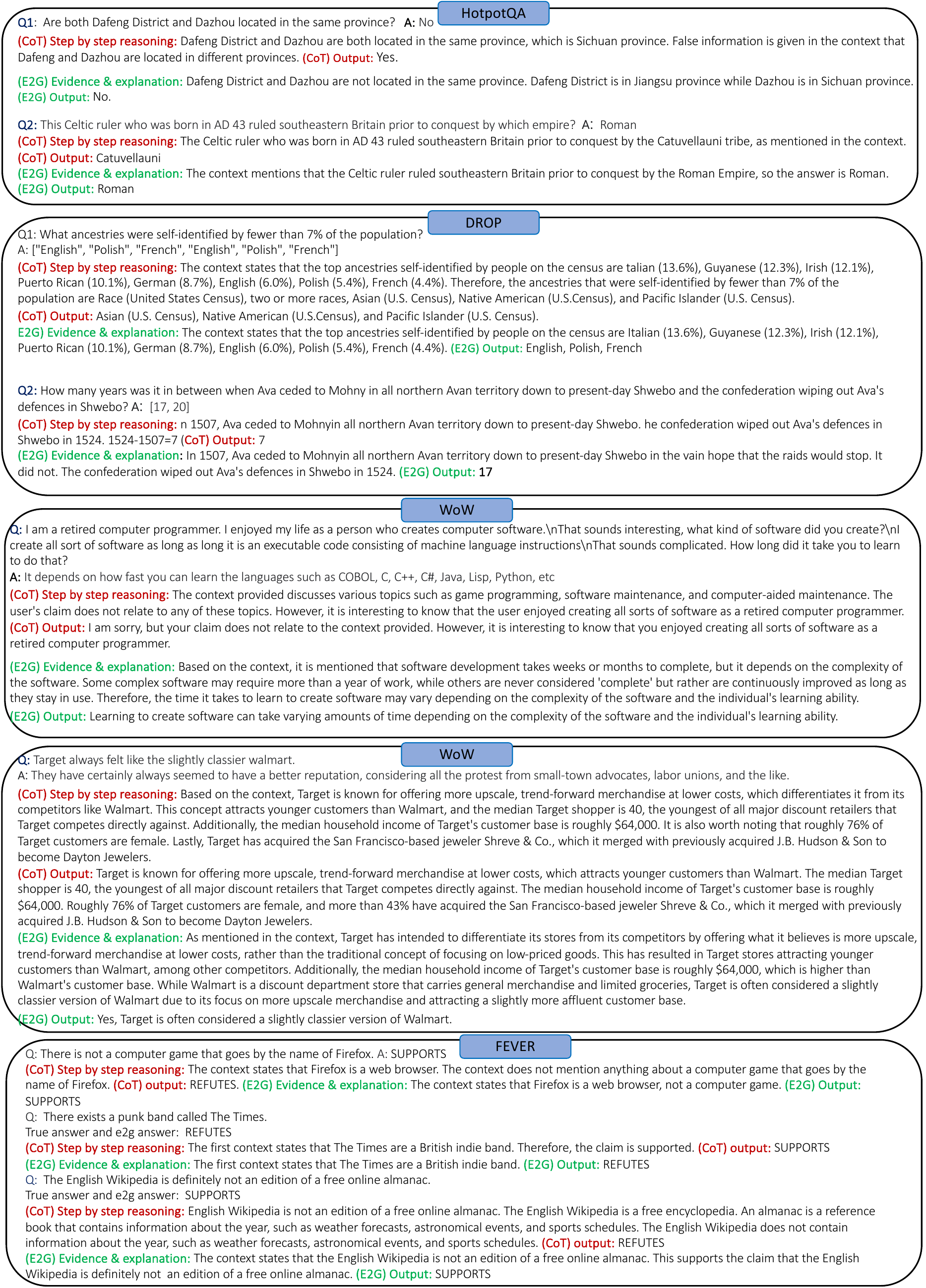}
    \caption{Example Model I/O (E2G vs CoT)}
    \label{fig:appendix:example:e2g-cot}
\end{figure*}

\begin{figure*}
    \centering
    \includegraphics[width=\linewidth]{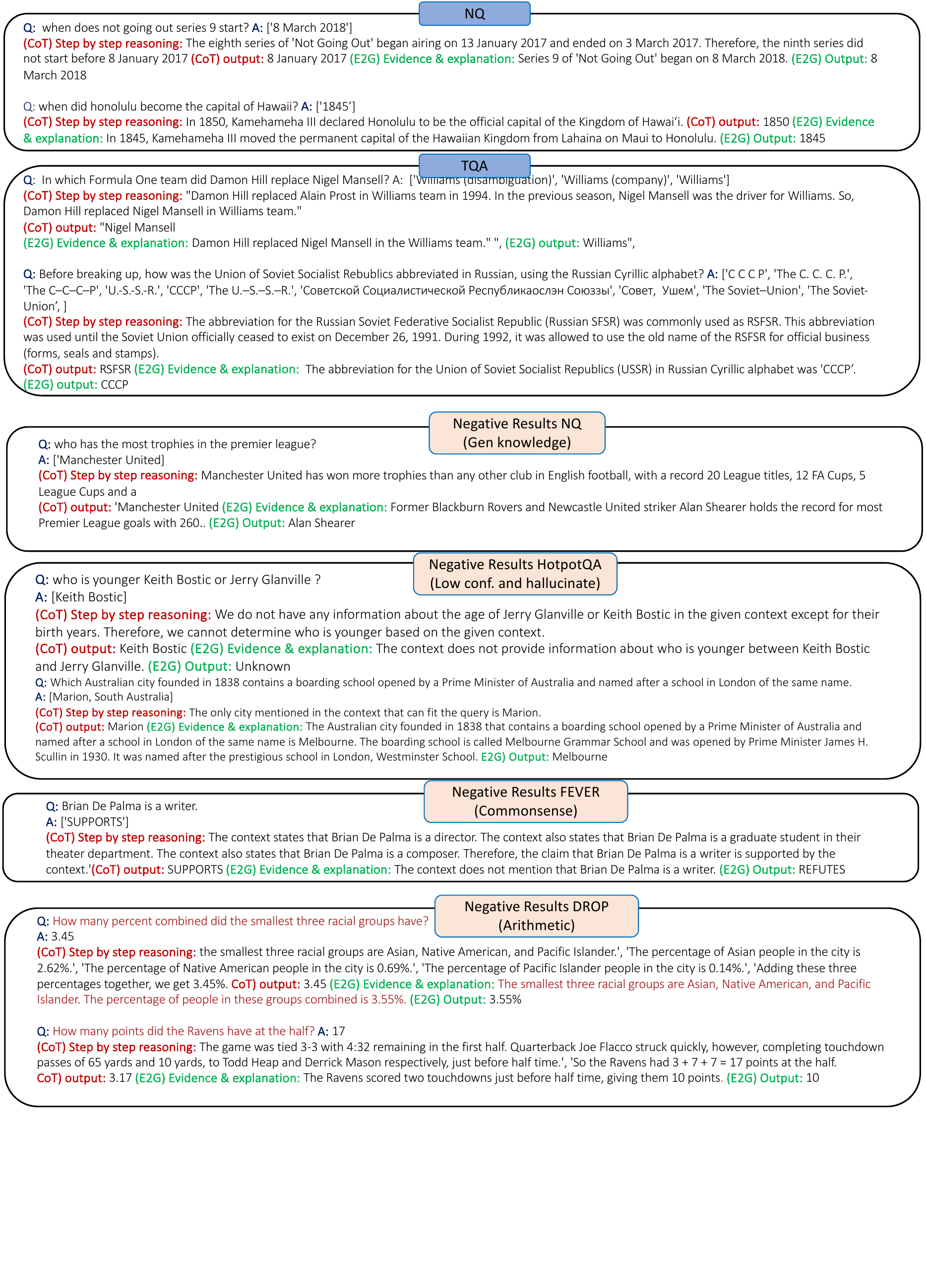}
    \caption{Example Model I/O (correct and incorrect outcome)}
    \label{fig:appendix:example:correct-incorrect}
\end{figure*}

\end{document}